\pdfoutput=1

\documentclass[11pt,table,dvipsnames]{article}

\usepackage{acl}
\usepackage{times}
\usepackage{latexsym}
\usepackage[T1]{fontenc}

\usepackage[utf8]{inputenc}

\usepackage{microtype}
\usepackage{latexsym}
\usepackage{dsfont}
\usepackage{booktabs} 
\usepackage{subfigure}
\usepackage{CJKutf8}
\usepackage{graphicx}
\usepackage{bigstrut}
\usepackage{multirow}
\usepackage{amsmath}
\usepackage{multirow, makecell, caption}
\usepackage{floatrow}
\newfloatcommand{capbtabbox}{table}[][\FBwidth]

\usepackage[normalem]{ulem}
\usepackage{amssymb}
\usepackage{amsfonts}
\usepackage{multicol}
\usepackage{tipa}
\usepackage{amsmath}
\usepackage{bm}
\usepackage[ruled,linesnumbered]{algorithm2e}

\usepackage{paralist}
\usepackage{IEEEtrantools}
\usepackage[switch]{lineno}
\usepackage{enumitem}
\usepackage{multirow, makecell, caption}
\usepackage{arydshln}
\usepackage{verbatim}
\usepackage[normalem]{ulem}
\usepackage{amssymb}
\usepackage{amsfonts}
\usepackage{multicol}

\usepackage{csquotes}
\usepackage{xspace}
\usepackage{paralist}
\usepackage{mdwlist}
\usepackage{subfigure}
\usepackage{makecell}
\usepackage{array}
\usepackage{enumitem}
\usepackage{bm}
\usepackage{colortbl}
\usepackage{dsfont}
\usepackage{url}
\usepackage{booktabs} 
\usepackage{graphicx}
\usepackage{tabularx}
\usepackage{amsmath}
\usepackage{bbm}
\usepackage{tikz}
\usepackage{pgfplots}
\newcommand{\method}{\textsc{KPCE}\xspace}
\newcommand{\cls}{\texttt{[CLS]}\xspace}
\newcommand{\sep}{\texttt{[SEP]}\xspace}

\newcommand{\ie}{\textit{i.e.}\xspace}
\newcommand{\eg}{\textit{e.g.}\xspace}
\newcommand{\wo}{\textit{w/o}\xspace}

\makeatletter
\def\adl@drawiv#1#2#3{%
        \hskip.5\tabcolsep
        \xleaders#3{#2.5\@tempdimb #1{1}#2.5\@tempdimb}%
                #2\z@ plus1fil minus1fil\relax
        \hskip.5\tabcolsep}
\newcommand{\cdashlinelr}[1]{%
  \noalign{\vskip\aboverulesep
           \global\let\@dashdrawstore\adl@draw
           \global\let\adl@draw\adl@drawiv}
  \cdashline{#1}
  \noalign{\global\let\adl@draw\@dashdrawstore
           \vskip\belowrulesep}}
\makeatother

\title{Causality-aware Concept Extraction \\based on Knowledge-guided Prompting}

\author{Siyu Yuan\textsuperscript{\rm $\heartsuit$},
Deqing Yang\textsuperscript{\rm $\heartsuit$}\thanks{~~Corresponding authors.},\\
\bf
Jinxi Liu\textsuperscript{\rm $\heartsuit$},
Shuyu Tian\textsuperscript{\rm $\heartsuit$},
Jiaqing Liang\textsuperscript{\rm $\heartsuit$},
Yanghua Xiao\textsuperscript{\rm $\spadesuit\clubsuit$}\footnotemark[1], Rui	Xie\textsuperscript{\rm $\diamondsuit$}\\
\textsuperscript{\rm $\heartsuit$}School of Data Science, Fudan University, Shanghai, China\\
\textsuperscript{\rm $\spadesuit$}Shanghai Key Laboratory of Data Science, School of Computer Science, Fudan University\\
\textsuperscript{\rm $\clubsuit$}Fudan-Aishu Cognitive Intelligence Joint Research Center, 
\textsuperscript{\rm $\diamondsuit$}Meituan, Beijing, China\\
\texttt{\{syyuan21,jxliu22, sytian21\}@m.fudan.edu.cn}, \\
\texttt{\{yangdeqing,liangjiaqing,shawyh\}@fudan.edu.cn}, 
\texttt{rui.xie@meituan.com}
}

\begin{document}

\maketitle

\begin{abstract}
Concepts benefit natural language understanding but are far from complete in existing knowledge graphs (KGs). 
Recently, pre-trained language models (PLMs) have been widely used in text-based concept extraction (CE). 
However, PLMs tend to mine the co-occurrence associations from massive corpus as pre-trained knowledge rather than the real causal effect between tokens.
As a result, the pre-trained knowledge confounds PLMs to extract biased concepts based on spurious co-occurrence correlations, inevitably resulting in low precision. 
In this paper, through the lens of a Structural Causal Model (SCM), we propose equipping the PLM-based extractor with a knowledge-guided prompt as an intervention to alleviate concept bias. 
The prompt adopts the topic of the given entity from the existing knowledge in KGs to mitigate the spurious co-occurrence correlations between entities and biased concepts. 
Our extensive experiments on representative multilingual KG datasets justify that our proposed prompt can effectively alleviate concept bias and improve the performance of PLM-based CE models.
The code has been released on \url{https://github.com/siyuyuan/KPCE}.
\end{abstract}

\section{Introduction}
\label{sec:intro}

The concepts in knowledge graphs (KGs) enable machines to understand natural languages better, and thus benefit many downstream tasks, such as question answering~\cite{han2020meta}, commonsense reasoning~\cite{zhong2021care} and entity typing~\cite{yuan-etal-2022-generative-entity}. 
However, the concepts, especially the fine-grained ones, in existing KGs still need to be completed. 
For example, in the widely used Chinese KG \textit{CN-DBpedia}~\cite{CN-DBpedia}, there are nearly 17 million entities but only 0.27 million concepts in total, and more than 20\% entities even have no concepts.
Although \textit{Probase}~\cite{Probase} is a large-scale English KG, the fine-grained concepts with two or more modifiers in it only account for 30\%~\cite{li2021towards}.
We focus on extracting multi-grained concepts from texts to complete existing KGs. 

Most of the existing text-based concept acquisition approaches adopt the extraction scheme, which can be divided into two categories:
\begin{inparaenum}[\it 1)]
    \item pattern-matching approaches~\cite{DBpedia,Probase,CN-DBpedia}, which can obtain high-quality concepts but only have low recall due to poor generalization;
    \item learning-based approaches~\cite{luo2020alicoco,ji2020fully,yuan2021large}, which employ pre-trained language models (PLMs) fine-tuned with labeled data to extract concepts.
\end{inparaenum}

\begin{figure}[t]
	\centering
	\includegraphics[width=0.95\columnwidth]{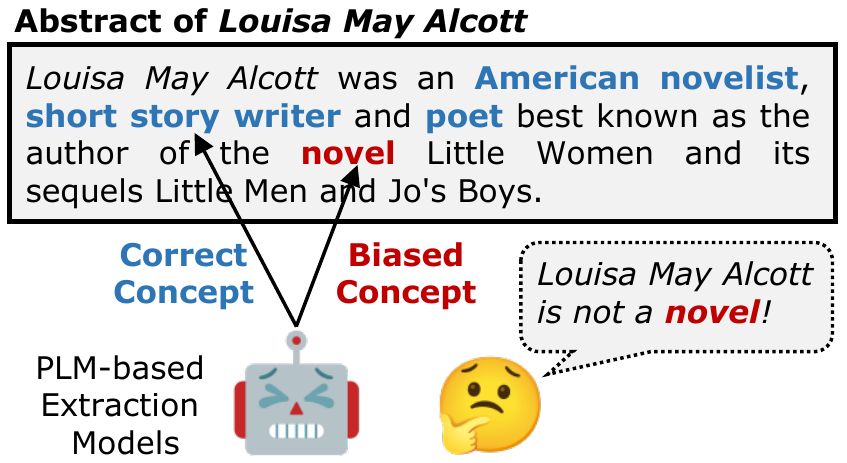} 
	\caption{The example of concept bias. The PLM-based CE models are biased to extract \textit{novel} mistakenly as the concept of \textit{Louisa May Alcott} from the text.}
	\label{fig:EB}
\end{figure}

However, an unignorable drawback of these learning-based approaches based on PLMs is \textbf{concept bias}. 
Concept bias means the concepts are extracted based on their contextual (co-occurrence) associations rather than the real causal effect between the entities and concepts, resulting in low extraction precision.
For example, in Figure~\ref{fig:EB}, PLMs tend to extract \textit{novel} and \textit{writer} together as concepts for the entity \textit{Louisa May Alcott} even if we explicitly input the entity \textit{Louisa May Alcott} to the model.
Previous work demonstrates that causal inference is a promising technique for bias analysis~\cite{lu2022neuro}.
To analyze the reasons behind concept bias, we devise a Structural Causal Model (SCM)~\cite{pearl2009causality} to investigate the causal effect in the PLM-based concept extraction (CE) system, and show that pre-trained knowledge in PLMs confounds PLMs to extract biased concepts.
During the pre-training, the entities and biased concepts (\eg, \textit{Louisa May Alcott} and \textit{novel}) often co-occur in many texts.
Thus, PLMs tend to mine statistical associations from a massive corpus rather than the real causal effect between them~\cite{li-etal-2022-pre}, which induces spurious co-occurrence correlations between entities (\ie, \textit{Louisa May Alcott}) and biased concepts (\ie, \textit{novel}).
Since we cannot directly observe the prior distribution of pre-trained knowledge, the backdoor adjustment is intractable for our problem~\cite{pearl2009causality}.
Alternatively, the frontdoor adjustment~\cite{peters2017elements} can apply a mediator as an intervention to mitigate bias.

In this paper, we adopt language prompting~\cite{gao2020making,li2021prefix} as a mediator for the frontdoor adjustment to handle concept bias.
We propose a novel \textbf{C}oncept \textbf{E}xtraction framework with \textbf{K}nowledge-guided \textbf{P}rompt, namely \textbf{\method} to extract concepts for given entities from text.
Specifically, we construct a knowledge-guided prompt by obtaining the topic of the given entity (\eg, \textit{person} for \textit{Louisa May Alcott}) from the knowledge in the existing KGs.
Our proposed knowledge-guided prompt is independent of pre-trained knowledge and fulfills the frontdoor criterion.
Thus, it can be used as a mediator to guide PLMs to focus on the right cause and alleviate spurious correlations.
Although adopting our knowledge-guided prompt to construct the mediator is straightforward, it has been proven effective in addressing concept bias and improving the extraction performance of PLM-based extractors in the CE task.

In summary, our contributions include:
\begin{inparaenum}[\it 1)]
    \item To the best of our knowledge, we are the first to identify the concept bias problem in the PLM-based CE system.
    \item We define a Structural Causal Model to analyze the concept bias from a causal perspective and propose adopting a knowledge-guided prompt as a mediator to alleviate the bias via frontdoor adjustment.
    \item Experimental results demonstrate the effectiveness of the proposed knowledge-guided prompt, which significantly mitigates the bias and achieves a new state-of-the-art for CE task. 
\end{inparaenum}

\section{Related Work}
\label{sec:relate}
\paragraph{Concept Acquisition}
Most of the existing text-based concept acquisition approaches adopt the extraction scheme, which can be divided into two categories:
\begin{inparaenum}[\it 1)]
    \item \textit{Pattern-matching Approaches}: 
    extract concepts from free texts with hand-crafted patterns~\cite{DBpedia,Probase,CN-DBpedia}. Although they can obtain high-quality concepts, they have low recall due to their poor generalization ability;
    \item \textit{Learning-based Approaches}: 
    mostly employ the PLM-based extraction models from other extraction tasks, such as the Named Entity Recognition (NER) models~\cite{li2020character,luo2021privacy,lange2022clin} and Information Extraction models~\cite{ijcai2021p200,yuan2021large} in the CE task. Although they can extract many concepts from a large corpus, the concept bias cannot be well handled. 
\end{inparaenum}

\paragraph{Causality for Language Processing}
Several recent work studies causal inference combined with language models for natural language processing (NLP)~\cite{scholkopf2022causality}, such as controllable text generation~\cite{hu2021a,goyal-etal-2022-cam} and counterfactual reasoning~\cite{Chen_Gan_Cheng_Zhou_Xiao_Li_2022,paranjape-etal-2022-retrieval}.
In addition, causal inference can recognize spurious correlations via Structural Causal Model (SCM)~\cite{pearl2009causality} for bias analysis and eliminate biases using causal intervention techniques~\cite{weber-etal-2020-causal,lu2022neuro}.
Therefore, there are also studies showing that causal inference is a promising technique to identify undesirable biases in the NLP dataset~\cite{feder-etal-2022-causal} pre-trained language models (PLMs)~\cite{li-etal-2022-pre}.
In this paper, we adopt causal inference to identify, understand, and alleviate concept bias in concept extraction.

\paragraph{Language Prompting}
Language prompting can distill knowledge from PLMs to improve the model performance in the downstream task. 
Language prompt construction methods can be divided into two categories~\cite{liu2021pre}:
\begin{inparaenum}[\it 1)]
    \item \textit{Hand-crafted Prompts}, 
    which are created manually based on human insights into the tasks~\cite{brown2020language,schick2020few,schick2020exploiting}. 
    Although they obtain high-quality results, how to construct optimal prompts for a certain downstream task is an intractable challenge;
    \item \textit{Automated Constructed Prompts},
    which are generated automatically from natural language phrases~\cite{jiang2020can,yuan2021bartscore} or vector space~\cite{li2021prefix,liu2021gpt}. 
\end{inparaenum}
Although previous work analyzes the prompt from a causal perspective~\cite{cao-etal-2022-prompt}, relatively little attention has been paid to adopting the prompt to alleviate the bias in the downstream task.

\section{Concept Bias Analysis}
\label{sec:Analysis}

In this section, we first formally define our task. Then we investigate the concept bias issued by PLMs in empirical studies. Finally, we devise a Structural Causal Model (SCM) to analyze the bias and alleviate it via causal inference.

\subsection{Preliminary}\label{subsec:Preliminary}
\paragraph{Task Definition}
Our CE task addressed in this paper can be formulated as follows. 
Given an entity $E=\{e_1, e_2, \cdots, e_{|{E}|}\}$ and its relevant text $T =\{t_1, t_2, \cdots, t_{|{T}|}\}$ where $e_i$ (or $t_i$) is a word token, our framework aims to extract one or multiple spans from $T$ as the concept(s) of ${E}$.

\paragraph{Data Selection}
It must guarantee that the given text contains concepts. 
The abstract text of an entity expresses the concepts of the entity explicitly, which can be obtained from online encyclopedias or knowledge bases.
In this paper, we take the abstract text of an entity as its relevant text $T$. 
The details of dataset construction will be introduced in $\mathsection$~\ref{subsec:dataset}.
Since we aim to extract concepts from $T$ for $E$, it is reasonable to concatenate $E$ and $T$ to form the input text $X =\{E,T\}$.
\begin{figure}[t]
	\centering
	\includegraphics[width=\columnwidth]{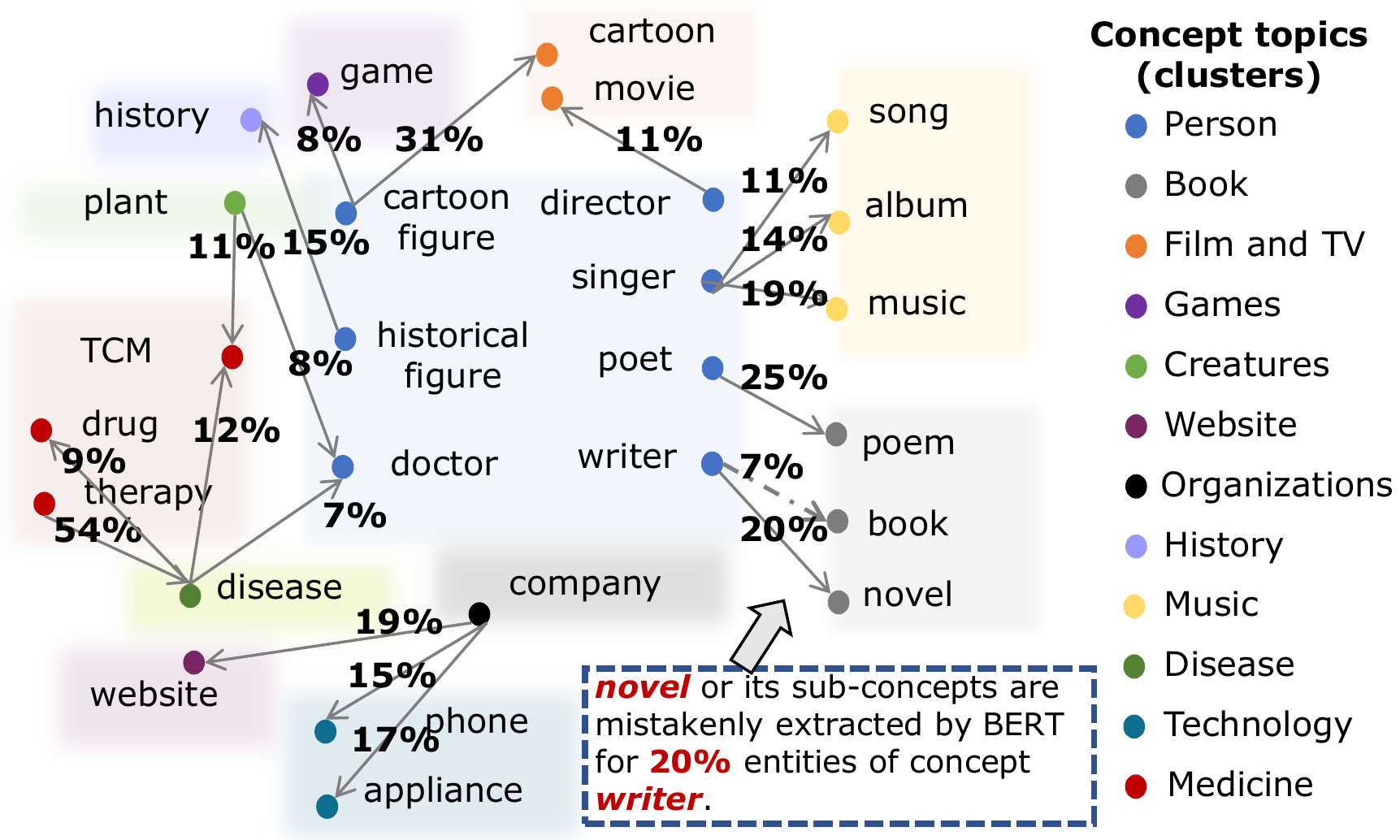}
	\caption{Concept bias map for the entities of popular concepts in CN-DBpedia (better viewed in color).}
	\label{fig:drift_pic}
\end{figure}

\subsection{Empirical Studies on Concept Bias}\label{subsec:Empirical_Studies}
To demonstrate the presence of concept bias, we conduct empirical studies on the CN-DBpedia dataset~\cite{CN-DBpedia}.
First, we randomly sample 1 million entities with their concepts from CN-DBpedia, and select the top 100 concepts with the most entities as the \emph{typical concept} set.
Then we randomly select 100 entities with their abstracts for each typical concept to construct the input texts and run a BERT-based extractor to extract concepts. 
Details of the extraction process will be introduced in $\mathsection$~\ref{subsec:extractor}.
We invite volunteers to assess whether the extracted concepts are biased. 
To quantify the degree of concept bias, we calculate the \textit{bias rate} of concept A to another concept B.
The bias rate is defined as the number of entities of A for which B or the sub-concepts of B are mistakenly extracted by the extractor, divided by the total number of entities of A.

The bias rates among 26 typical concepts are shown in Figure~\ref{fig:drift_pic}, where the concepts (dots) of the same topic are clustered in one rectangle. 
The construction of concept topics will be introduced in $\mathsection$~\ref{subsec:topic}. 
From the figure, we can conclude that concept bias is widespread in the PLM-based CE system and negatively affects the quality of the results.
Previous studies have proven that causal inference can analyze bias via SCM and eliminate bias with causal intervention techniques~\cite{cao-etal-2022-prompt}.
Next, we will analyze concept bias from a causal perspective.

\begin{figure}[t]
	\centering
	\includegraphics[width=0.8\columnwidth]{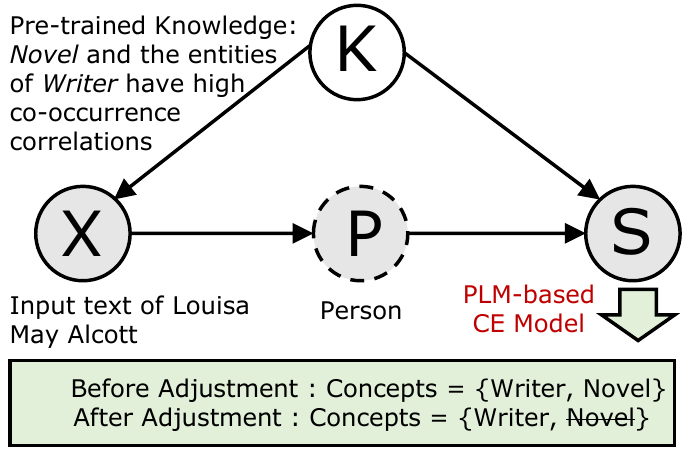}
	\caption{The proposed structural causal model (SCM). A hollow circle indicates the variable is latent, and a shaded circle indicates the variable is observed. 
Without causal intervention, the PLM-based CE model
extract \textit{Novel} due to the spurious correlation between the entities of \textit{Writer} and \textit{Novel} caused by the confounding variable $K$. The constructed
mediating variable $P$ can block the backdoor paths for $X \rightarrow S$ (opened
by $K$) and help the model only extract the unbiased concept \textit{Writer}. 
 }
	\label{fig:SCM}
\end{figure}

\subsection{The Causal Framework for Concept Bias Analysis}\label{sec:SCM_analysis}
\paragraph{The Structural Causal Model}
We devise a Structural Causal Model (SCM) to identify the causal effect between the input text $X$ of a given entity $E$ and the concept span $S$ that can be extracted from ${X}$. 
As shown in Figure~\ref{fig:SCM}, our CE task aims to extract one or multiple spans $S$ from ${X}$ as the concept(s) of ${E}$ where the causal effect can be denoted as $X \rightarrow S$.

%

During the pre-training, the contextual embedding of one token depends on the ones that frequently appear nearby in the corpus.
We extrapolate that the high co-occurrence between the entities of true concepts (\eg, \textit{writer}) and biased concepts (\eg, \textit{novel}) in the pre-trained knowledge induces spurious correlations between entities (\eg, \textit{Louisa May Alcott}) and biased concepts (\eg, \textit{novel}).
Therefore, the PLM-based CE models can mistakenly extract biased concepts even if the entity is explicitly mentioned in $X$.
The experiments in $\mathsection$~\ref{subsec:analysis} also prove our rationale.
Based on the foregoing analysis, we define the pre-trained knowledge ${K}$ from PLM-based extraction models as a confounder.

We cannot directly observe the latent space of the PLMs, and thus the backdoor adjustment~\cite{pearl2009causality} is not applicable in our case.
Alternatively, we adopt the frontdoor adjustment~\cite{peters2017elements} and design a mediator to mitigate the concept bias.

\paragraph{Causal Intervention}
To mitigate the concept bias, we construct a prompt $P$ as a mediator for $X \rightarrow S$, and then the frontdoor adjustment can apply do-operation.

Specifically, to make the PLMs attend to the right cause and alleviate spurious co-occurrence correlation (\eg, \textit{novel} and \textit{Louisa May Alcott}), we assign a topic as a knowledge-guided prompt $P$ (\ie, \textit{person}) to the input text $X$ (The detailed operation is elaborated in $\mathsection$~\ref{subsec:topic}). 
The topics obtained from KGs are independent of pre-trained knowledge, and thus $P$ fulfills the frontdoor criterion.

For the causal effect $X \rightarrow P$, we can observe that $X \rightarrow P \rightarrow S \leftarrow K$ is a collider that blocks the association between $P$ and $K$, and no backdoor path is available for $X \rightarrow P$.
Therefore, we can directly rely on the conditional probability after applying the do-operator for $X$:
\begin{equation}\label{eq:step1}
	P(P=p|do(X=x))=P(P=p|X=x).
\end{equation}

Next, for the causal effect $P \rightarrow S$, $P \leftarrow X \leftarrow K \rightarrow S$ is a backdoor path from $P$ to $S$, which we need to cut off. 
Since $K$ is an unobserved variable, we can block the backdoor path through $X$:
\begin{equation}\label{eq:step2}
	P(S|do(P))=\sum_x P(S|P,X=x)P(X=x).
\end{equation}
Therefore, the underlying causal mechanism of our CE task is a combination of Eq.\ref{eq:step1} and Eq.\ref{eq:step2}, which can be formulated as:
\begin{align}\label{eq:final}
	&P(S|do(X))\notag\\
	&=\sum_{p} P(S|p,do(X))P(p|do(X))\notag\\
	&=\sum_{p} P(S|do(P),do(X))P(p|do(X))\notag\\
	&=\sum_{p} P(S|do(P))P(p|do(X)).
\end{align}
The theoretical details of the frontdoor adjustment are introduced in Appendix~\ref{sec:theoretical_details}.

We make the assumption of strong ignorability, \ie, there is only one confounder $K$ between $X$ and $S$. 
One assumption of the frontdoor criterion is that the only way the input text $X$ influences $S$ is through the mediator $P$. 
Thus, $X \rightarrow P \rightarrow S$ must be the only path. Otherwise, the front-door adjustment cannot stand.
Notice that $K$ already represents all the knowledge from pre-trained data in PLMs. 
Therefore, it is reasonable to use the strong ignorability assumption that it already includes all possible confounders.

Through the frontdoor adjustment, we can block the backdoor path from input text to concepts and alleviate spurious correlation caused by the confounder, \ie, pre-trained knowledge.
In practice, we can train a topic classifier to estimate Eq.\ref{eq:step1} ($\mathsection$~\ref{subsec:topic}) and train a concept extractor on our training data to estimate Eq.\ref{eq:step2} ($\mathsection$~\ref{subsec:extractor}).
Next, we will introduce the implementation of the frontdoor adjustment in detail.

\section{Methodology}
\label{sec:Methodology}

\begin{figure}[t]
	\centering
	\includegraphics[width=\columnwidth]{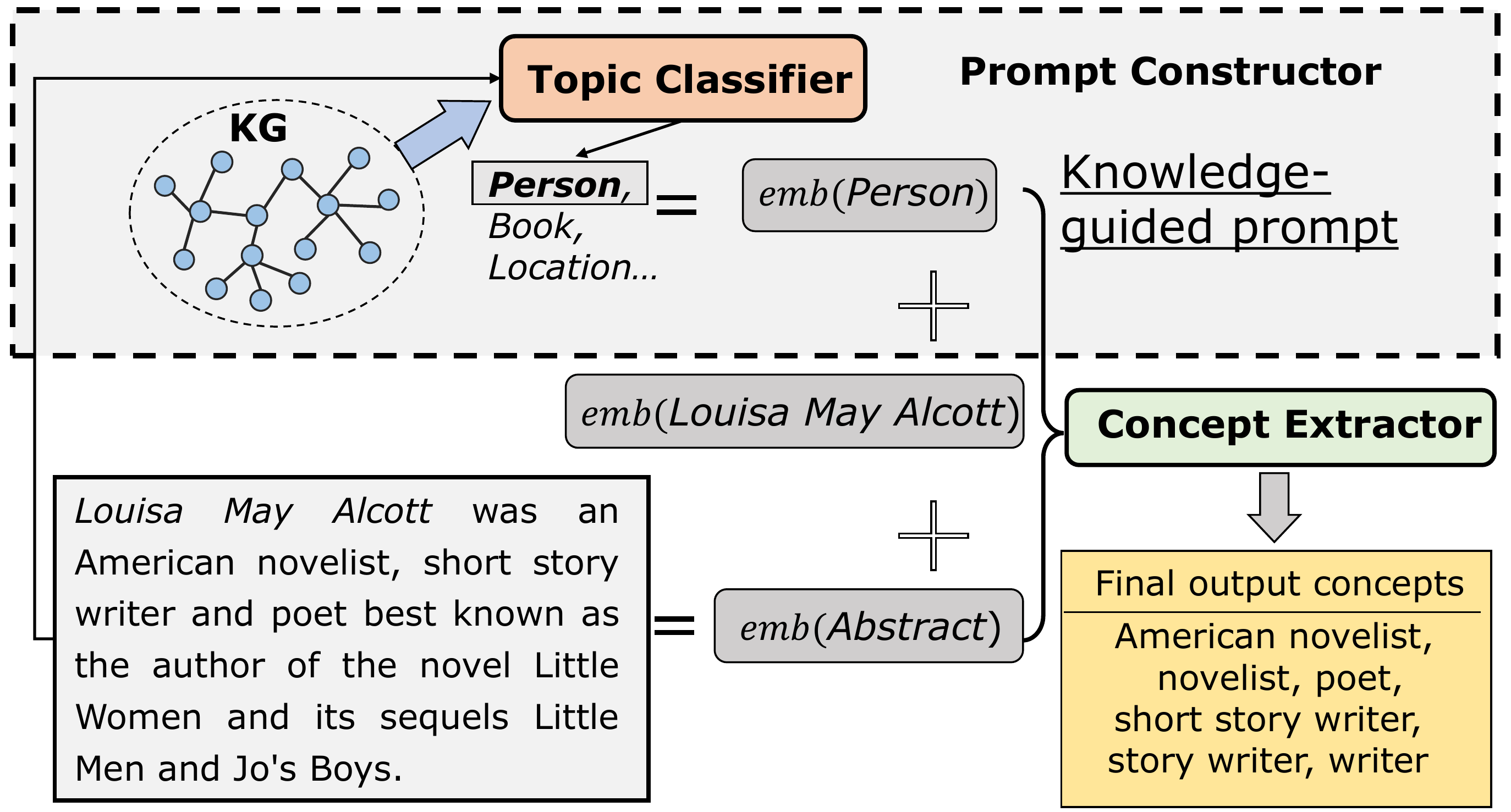}
	\caption{The overview of our CE framework.}
	\label{fig:Model}
\end{figure}

In this section, we present our CE framework \method and discuss how to perform prompting to alleviate concept bias.
The overall framework of \method is illustrated in Figure~\ref{fig:Model}, which consists of two major modules:
\begin{inparaenum}[\it 1)]
    \item \textit{Prompt Constructor}: assigns the topic obtained from KGs for entities as a knowledge-guided prompt to estimate Eq.\ref{eq:step1};
    \item \textit{Concept Extractor}: trains a BERT-based extractor with the constructed prompt to estimate Eq.\ref{eq:step2} and extract multi-grained concepts from the input text.
\end{inparaenum}
Next, we will introduce the two modules of \method.

\subsection{Prompt Constructor}\label{subsec:topic}
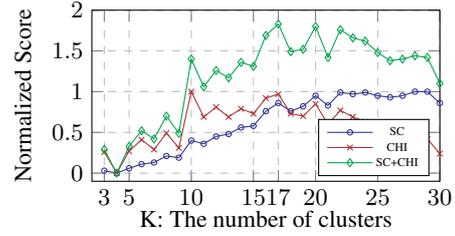
\begin{figure}[t]
    \centering
\pgfplotsset{width=0.8\linewidth,height=0.5\linewidth,compat=1.16}
\footnotesize
\begin{tikzpicture}
\begin{axis}[
    xlabel={K: The number of clusters},
    ylabel={Normalized Score},
    xmin=2, xmax=30,
    ymin=-0.1, ymax=2,
    xtick={3, 5, 10, 15, 17, 20, 25, 30},
    ytick={0, 0.5, 1.0, 1.5, 2.0},
    legend pos=south east,
    ymajorgrids=true,
    xmajorgrids=true,
    grid style=dashed,
    x label style={at={(axis description cs:0.5,-0.125)},anchor=north},
    y label style={at={(axis description cs:-0.15,0.5)},anchor=south},
    legend style={nodes={scale=0.5, transform shape}}
]
\addplot[
    color=Blue,
    mark=o,
    mark size=1pt,
    ]
    coordinates {
    (3, 0.03)
(4, 0.0)
(5, 0.06)
(6, 0.11)
(7, 0.13)
(8, 0.21)
(9, 0.19)
(10, 0.4)
(11, 0.36)
(12, 0.45)
(13, 0.48)
(14, 0.56)
(15, 0.58)
(16, 0.76)
(17, 0.86)
(18, 0.76)
(19, 0.82)
(20, 0.95)
(21, 0.83)
(22, 0.99)
(23, 0.97)
(24, 0.99)
(25, 0.95)
(26, 0.93)
(27, 0.95)
(28, 1.0)
(29, 1.0)
(30, 0.86)
    };
    \addlegendentry{SC}
    
\addplot[
    color=Maroon,
    mark=x,
    mark size=1.5pt,
    ]
    coordinates {
    (3, 0.26)
(4, 0.0)
(5, 0.27)
(6, 0.41)
(7, 0.29)
(8, 0.49)
(9, 0.31)
(10, 1.0)
(11, 0.69)
(12, 0.81)
(13, 0.69)
(14, 0.79)
(15, 0.73)
(16, 0.92)
(17, 0.97)
(18, 0.73)
(19, 0.7)
(20, 0.85)
(21, 0.6)
(22, 0.77)
(23, 0.69)
(24, 0.63)
(25, 0.53)
(26, 0.45)
(27, 0.45)
(28, 0.44)
(29, 0.42)
(30, 0.24)
    };
    \addlegendentry{CHI}

\addplot[
    color=Green,
    mark=diamond,
    mark size=1.5pt,
    ]
    coordinates {
    (3, 0.29)
(4, 0.0)
(5, 0.33)
(6, 0.52)
(7, 0.42)
(8, 0.7)
(9, 0.49)
(10, 1.4)
(11, 1.06)
(12, 1.26)
(13, 1.17)
(14, 1.36)
(15, 1.31)
(16, 1.69)
(17, 1.83)
(18, 1.49)
(19, 1.52)
(20, 1.80)
(21, 1.42)
(22, 1.76)
(23, 1.66)
(24, 1.62)
(25, 1.48)
(26, 1.38)
(27, 1.4)
(28, 1.44)
(29, 1.42)
(30, 1.1)
    };
    \addlegendentry{SC+CHI}
    
\end{axis}
\end{tikzpicture}
    \caption{The scores of Silhouette Coefficient (SC) and Calinski Harabaz Index (CHI) under different cluster numbers. The scores are normalized with feature scaling for a fair comparison.}
    \label{fig:cluster}
\end{figure}

\paragraph{Knowledge-guided Prompt Construction}
To reduce the concept bias, we use the topic of a given entity as a knowledge-guided prompt, which is identified based on the external knowledge of the existing KGs.
Take \emph{CN-DBpedia}~\cite{CN-DBpedia} as an example~\footnote{In fact, the concepts of CN-DBpedia are inherited from Probase, so the typical topics are the same for CN-DBpedia and Probase.}. 
We randomly sample one million entities from this KG and obtain their existing concepts. 
Then, we select the top 100 concepts having the most entities to constitute the \emph{typical concept} set, which can cover more than 99.80\% entities in the KG. 
Next, we use spectral clustering~\cite{von2007tutorial} with the adaptive K-means~\cite{bhatia2004adaptive} algorithm to cluster these typical concepts into several groups, each of which corresponds to a topic. 
To achieve the spectral clustering, we use the following overlap coefficient~\cite{vijaymeena2016survey} to measure the similarity between two concepts,
\begin{equation}\label{eq:coefficient}
	Overlap(c_{1},c_{2}) =\frac{|ent(c_{1}) \cap ent(c_{2})|}{min(|ent(c_{1})|, |ent(c_{2})|)+\delta}
\end{equation}
where $ent(c_{1})$ and $ent(c_{2})$ are the entity sets of concept $c_1$ and concept $c_2$, respectively. 
We then construct a similarity matrix of typical concepts to achieve spectral clustering. 
To determine the best number of clusters, we calculate the Silhouette Coefficient (SC)~\cite{aranganayagi2007clustering} and the Calinski Harabaz Index (CHI)~\cite{maulik2002performance} from 3 to 30 clusters.
The scores are shown in Figure~\ref{fig:cluster}, from which we find that the best number of clusters is 17. 
As a result, we cluster the typical concepts into 17 groups and define a topic name for each group. 
The 17 typical topics and their corresponding concepts are listed in Appendix~\ref{sec:classifier}
\paragraph{Identifying Topic Prompt for Each Entity}
We adopt a topic classifier to assign the topic prompt to the input text $X$, which is one of the 17 typical topics in Table~\ref{tb:topic}.
To construct the training data, we randomly fetch 40,000 entities together with their abstract texts and existing concepts in the KG. 
According to the concept clustering results, we can assign each topic to the entities.
We adopt transformer encoder~\cite{NIPS2017_3f5ee243} followed by a two-layer perception (MLP)~\cite{gardner1998artificial} activated by ReLU, as our topic classifier~\footnote{We do not employ the PLM-based topic classifier since it will bring a direct path from $K$ to $P$ in Figure~\ref{fig:SCM}.}. 
We train the topic classifier to predict the topic prompt ${P}=\{p_1, p_2, \cdots, p_{|{P}|}\}$ for ${X}$, which is calculated as~\footnote{The detailed training operation of topic classifier can be found in Appendix~\ref{sec:classifier}}:
\begin{equation}\label{eq:topic}
	\begin{split}
		{P} = \mathop{\arg\max}\limits_{i}\big(P({P^i}|X)\big), 1\leq i \leq 17,
	\end{split}
\end{equation}
where ${P^i}$ is the i-th topic among the 17 typical topics.

In our experiments, the topic classifier achieves more than 97.8\% accuracy in 500 samples by human assessment. 
Through training the topic classifier, we can estimate Eq.\ref{eq:step1} to identify the causal effect $X \rightarrow P$.


\subsection{Concept Extractor}\label{subsec:extractor}
\paragraph{Prompt-based BERT}
The concept extractor is a BERT equipped with our proposed prompt followed by a pointer network~\cite{PN}.
The pointer network is adopted for extracting multi-grained concepts. 

We first concatenate the token sequence with the tokens of ${P}$ and ${X}$ to constitute the input, \ie, $\{\cls{P}\sep {X}\sep\}$, where $\cls$ and $\sep$ are the special tokens in BERT.
With multi-headed self-attention operations over the above input, the BERT outputs the final hidden state (matrix), \ie, $\mathbf{H}^{N_{L}}\in\mathbb{R}^{(|{P}|+|{X}|+3) \times d'}$ where $d'$ is the vector dimension and $N_{L}$ is the total number of layers. 
Then the pointer network predicts the probability of a token being the start position and the end position of the extracted span. 
We use $\mathbf{p}^{start},\mathbf{p}^{end}\in \mathbb{R}^{|{P}|+|{X}|+3}$ to denote the vectors storing the probabilities of all tokens to be the start position and end position, which are calculated as
\begin{equation}\label{eq:pn}
[\mathbf{p}^{start};\mathbf{p}^{end}] = \textrm{softmax}(\mathbf{H}^{N_{L}}\mathbf{W}+\mathbf{B})
\end{equation}
where $\mathbf{B}\in \mathbb{R}^{(|{P}|+|{X}|+3) \times 2}$ and $\mathbf{W}\in \mathbb{R}^{d' \times 2}$ and are both trainable parameters. 
We only consider the probabilities of the tokens in the abstract $T$. 
Given a span with $x_i$ and $x_j$ as the start token and the end token, its confidence score ${cs}_{ij} \in \mathbb{R}$ can be calculated as
\begin{equation}\label{eq:cs}
cs_{ij} =p^{start}_i+p^{end}_j.
\end{equation}
Accordingly, the model outputs a ranking list of candidate concepts (spans) with their confidence scores.
We only reserve the concepts with confidence scores bigger than the selection threshold.
An example to illustrate how to perform the pointer network is provided in Appendix~\ref{sec:PN}. 

During training, the concept extractor is fed with the input texts with topic prompts and outputs the probability (confidence scores) of the spans, and thus can estimate the causal effect $P \rightarrow S$ in Eq.\ref{eq:step2}.

\paragraph{Model Training}
We adopt the cross-entropy function $\mathop{CE}(\cdot)$ as the loss function of our model. 
Specifically, suppose that $\mathbf{y}_{start}\in \mathbb{N}^{|{P}|+|{X}|+3}$ (or $\mathbf{y}_{end}\in \mathbb{N}^{|{P}|+|{X}|+3}$) contains the real label (0 or 1) of each input token being the start (or end) position of a concept. 
Then, we have the following two training losses for the predictions:
\begin{align}\label{eq:loss1}
 \mathcal{L}_{start} &=\mathop{CE}(\mathbf{p}^{start},\mathbf{y}_{start}),\\
 \mathcal{L}_{end}&=\mathop{CE}(\mathbf{p}^{end},\mathbf{y}_{end}).
\end{align}
Then, the overall training loss is 
\begin{equation}\label{eq:loss2}
\mathcal{L}=\alpha \mathcal{L}_{start}+(1-\alpha)\mathcal{L}_{end}
\end{equation}
where $\alpha \in (0,1)$ is the control parameter. 
We use Adam~\cite{Adam} to optimize $\mathcal{L}$.


\section{Experiments}
\label{sec:Experiments}

\subsection{Datasets}\label{subsec:dataset}
\paragraph{CN-DBpedia}
From the latest version of Chinese KG CN-DBpedia~\cite{CN-DBpedia} and Wikipedia, we randomly sample 100,000 instances to construct our sample pool. 
Each instance in the sample pool consists of an entity with its concept and abstract text~\footnote{If one entity has multiple concepts, we randomly select one as the golden label.}. 
Then, we sample 500 instances from the pool as our test set and divide the rest of the instances into the training set and validation set according to 9:1. 

\paragraph{Probase}
We obtain the English sample pool of 50,000 instances from Probase~\cite{Probase} and Wikipedia. 
The training, validation and test set construction are the same as the Chinese dataset.

\subsection{Evaluation Metrics}
We compare \method with seven baselines, including a pattern-matching approach \ie, Hearst pattern.
Detailed information on baselines and some experiment settings is shown in Appendix \ref{sec:base} and \ref{sec:set}.
Some extracted concepts do not exist in the KG, and cannot be assessed automatically. 
Therefore, we invite the annotators to assess whether the extracted concepts are correct. 
The annotation detail is shown in Appendix~\ref{sec:Human_Assessment}.

Please note that the extracted concepts may already have existed in the KG for the given entity, which we denote as \textit{EC}s (existing concepts).
However, our work expects to extract correct but new concepts (that do not exist in the KG) to complete the KGs, which we denote as \textit{NC}s (new concepts).
Therefore, we record the number of new concepts (NC~\#) and display the ratio of correct concepts (ECs and NCs) as precision (Prec.).
Since it is difficult to know all the correct concepts in the input text, we report the relative recall (Recall$_R$). 
Specifically, suppose NCs~\# is the total number of new concepts extracted by all models. 
Then, the relative recall is calculated as NC~\# divided by NCs~\#~\footnote{Please note that NCs~\# is counted based on all models in one comparison. Therefore, Recall$_R$ can be different for one model when the compared models change.}.
Accordingly, the relative F1 (F1$_R$) can be calculated with Prec. and Recall$_R$. 
In addition, we also record the average length of new concepts (Len$_{NC}$) to investigate the effectiveness of the pointer network.

\subsection{Overall Performance}
We present the main results in Table~\ref{tab:overall}. 
Generally, we have the following findings: 

Our method outperforms previous baselines by large margins, including previous state-of-the-art (MRC-CE,~\citealp{yuan2021large}).
However, the pattern-based approach still beats the learning-based ones in precision, envisioning a room for improvement.
We find that \method achieves a more significant improvement in extracting new concepts, indicating that \method can be applied to achieve KG completion ($\mathsection$~\ref{subsec:Applications}).
We also compare \method with its ablated variant and the results show that adding a knowledge-guided prompt can guide BERT to achieve accurate CE results.


We notice that almost all models have higher extraction precision on the Chinese dataset than that on the English dataset.
This is because the modifiers are usually placed before nouns in Chinese syntactic structure, and thus it is easier to identify these modifiers and extract them with the coarse-grained concepts together to form the fine-grained ones. 
However, for the English dataset, not only adjectives but also subordinate clauses modify coarse-grained concepts, and thus identifying these modifiers is more difficult.

Compared with learning-based baselines, \method can extract more fine-grained concepts. 
Although the Hearst pattern can also extract fine-grained concepts, it cannot simultaneously extract multi-grained concepts when a coarse-grained concept term is the subsequence of another fine-grained concept term.
For example, in Figure~\ref{fig:Model}, if Hearst Pattern extracts \textit{American novelist} as a concept, it cannot extract \textit{novelist} simultaneously. 
\method solves this problem well with the aid of the pointer network and achieves a much higher recall.

\setlength\tabcolsep{4pt}
\begin{table}[t]
\small
  \centering
 \begin{tabular}{lcccccc}
 \toprule
 \textbf{Model}  & \textbf{NC \#}  & \textbf{Len$_{NC}$} & \textbf{Prec.}   & \textbf{Recall$_R$}  & \textbf{F1$_R$} \\
 \midrule
 \rowcolor[gray]{0.95}\multicolumn{7}{c}{\textit{Trained on CN-DBpedia}} \\
 Hearst  & 222   & \textbf{5.95}  & \textbf{95.24\%} &  21.66\%   & 35.29\% \\
 \cdashlinelr{1-6}
 FLAIR  & 64     & 3.09 & 95.71\%   & 6.24\% & 11.72\%   \\
 XLNet & 47     & 2.66 & 88.48\%   & 4.68\% & 8.90\% \\
 KVMN  & 254  & 4.03 & 64.45\%   & 26.02\%   & 37.08\%  \\
 XLM-R & 255   & 5.35 & 76.82\%   & 24.78\%   & 37.47\%  \\
 BBF  & 26     & 4.34 & 88.28\%   & 2.54\% & 4.93\% \\ 
 GACEN & 346   &3.58 & 84.89\%   & 36.73\% & 51.27\% \\
 MRC-CE & 323    & 5.33   & 92.12\%   & 31.51\% & 46.96\%  \\
 \cdashlinelr{1-6}
 \method  & \textbf{482}    & 5.52 & 94.20\%   & \textbf{44.38\%} & \textbf{60.33\%} \\
\ \ \ \ \wo \textit{P} & 338  & 5.21 &72.07\% & 34.05\% & 46.25\% \\
 \midrule
 \rowcolor[gray]{0.95}\multicolumn{7}{c}{\textit{Trained on Probose}} \\
 Hearst & 287   & \textbf{2.43}  & \textbf{89.04\%}   & 17.10\%   & 28.69\%   \\  
  \cdashlinelr{1-6}
 FLAIR & 140   & 1.68 & 84.31\%   & 7.73\% & 14.16\%   \\ 
 XLNet & 342   & 1.51 &79.30\% & 18.87\%   & 30.49\%   \\ 
  KVMN & 403   & 1.97 & 47.39\%   & 22.24\%   & 30.27\%   \\ 
  XLM-R   & 322  & 2.28 & 81.73\%   & 17.77\%   & 29.19\%   \\
  BBC   & 154  & 1.68 & 81.13\%   & 8.44\% & 15.30\%   \\ 
  GACEN & 486   &1.75 & 76.93\%   & 31.82\% & 45.02\% \\
  MRC-CE &598   &2.23 & 88.59\%   & 33.00\% & 48.09\% \\
  \cdashlinelr{1-6}
  \method  &   \textbf{752}   & 2.31 & 88.69\%   & \textbf{46.83\%} & \textbf{61.30\%} \\ 
    \ \ \ \ \wo \textit{P} & 691 & 2.26 & 78.64\% & 40.62\% & 53.57 \% \\
 \bottomrule
 \end{tabular}
  \caption{Concept extraction performance comparisons of 500 test samples. \wo \textit{P} is the ablation variants of \method without the knowledge-guided prompt (P)}
  \label{tab:overall}
\end{table}

\subsection{Analysis}\label{subsec:analysis}
In response to the motivations of \method, we conduct detailed analyses to further understand \method and why it works.

\setlength\tabcolsep{3pt}
\begin{table}[t]
    \small
    \centering
    \begin{tabular}{cccc}
    \toprule
        \textbf{Concept$_{O}$}&\textbf{Concept$_{B}$}&\textbf{\method$_{\wo P}$}&\textbf{\method}\\
    \midrule
    writer& book & 20\%& 7\%\\
    plant& doctor & 8\%& 0\%\\
    illness& medicine & 12\%& 6\%\\
    singer& music & 19\%& 2\%\\
    poem& poet & 25\%& 1\%\\
    \bottomrule
    \end{tabular}
    \caption{The bias rates issued by \method \wo \textit{P} and \method in five selected concepts. \textit{Concept$_{O}$} is the original concept and \textit{Concept$_{B}$} is the biased concept.}
    \label{tab:Extraction_Bias}
\end{table}

\begin{table}[t]
\centering
\small
\begin{tabular}{cccc}
\toprule
\multicolumn{4}{c}{\makecell[c]{
\textbf{Topic}: Technology. \textbf{Entity}: Korean alphabet. \\ \textbf{Abstract}: The Korean alphabet is a writing system for the \\ Korean language created by King Sejong the Great in 1443.}}\\
\midrule
\multicolumn{4}{c}{\textbf{Output Results}} \\
\hline
\multicolumn{2}{c}{\method$_{\wo P}$}  & \multicolumn{2}{c}{\method}\\ 
\hline
\multicolumn{1}{l}{\textbf{Span}}   & \multicolumn{1}{c}{\textbf{C.S.}}  & \multicolumn{1}{l}{\textbf{Span}} & \textbf{C.S.}  \\ \hline
\multicolumn{1}{l}{language}    & \multicolumn{1}{c}{0.238} & \multicolumn{1}{l}{system}   & 0.240 \\ 
\multicolumn{1}{l}{alphabet}   & \multicolumn{1}{c}{0.213} & \multicolumn{1}{l}{writing system}   & 0.219 \\ 
\multicolumn{1}{l}{system}   & \multicolumn{1}{c}{0.209} & \multicolumn{1}{l}{system for the Korean language}  & 0.130 \\ 
\bottomrule
\end{tabular}
\caption{A case to verify the effectiveness of the proposed prompts on addressing concept bias. We display an entity \textit{Korean alphabet} with its top-3 extracted spans and the confidence scores (denoted as C.S.)}\label{tb:Case_Study}
\end{table}

\paragraph{How does \method alleviate the concept bias?}
As mentioned in $\mathsection$~\ref{subsec:Empirical_Studies}, the concept bias occurs primarily among 26 concepts in CN-DBpedia. 
To justify that \method can alleviate concept bias with the aid of prompts, we randomly select five concepts and run \method with its ablated variant to extract concepts for 100 entities randomly selected from each of the five concepts.
Then we calculate the bias rates of each concept, and the results in Table~\ref{tab:Extraction_Bias} show that \method has a much lower bias rate than the vanilla BERT-based concept extractor. 
Thus, the knowledge-guided prompt can significantly mitigate the concept bias.

Furthermore, a case study for an entity \textit{Korean alphabet} is shown in Table \ref{tb:Case_Study}. 
We find that the proposed prompts can mitigate the spurious co-occurrence correlation between entities and biased concepts by decreasing the confidence scores of biased concepts (\ie, \textit{language} and \textit{alphabet}) and increasing the scores of correct concepts (\ie, \textit{system} and \textit{writing system}).
Thus, the knowledge-guided prompt can significantly alleviate the concept bias and result in more accurate CE results.
\begin{figure}[t]
	\centering
	\includegraphics[width=\columnwidth]{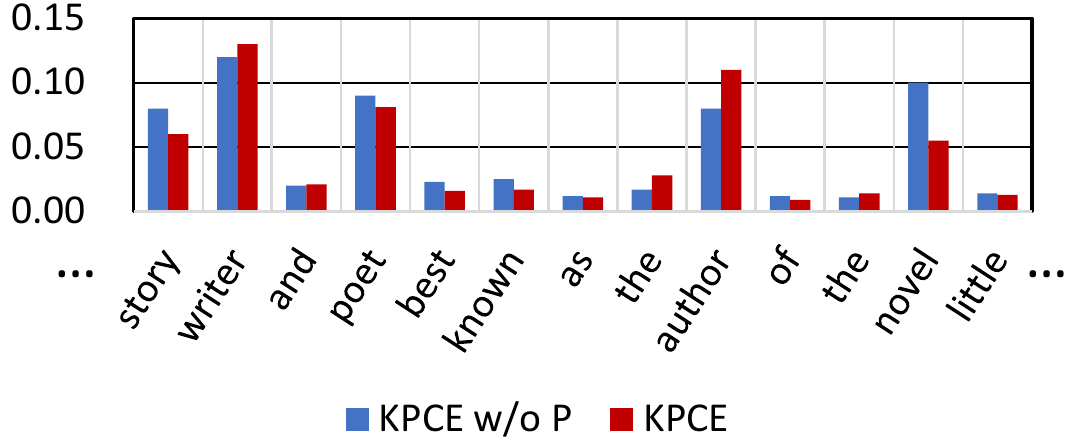}
	\caption{Visualization of the attention distribution of \cls to other tokens.}
	\label{fig:attention_dist}
\end{figure}

\paragraph{How does the prompt affect the spurious co-occurrence correlations?}
To explore the rationale behind the prompt-based mediator, we focus on the attention distribution for the special token \cls, since it is an aggregate representation of the sequence and can capture the sentence-level semantic meaning~\cite{devlin-etal-2019-bert,chang2022multi}.
Following previous work~\cite{clark-etal-2019-bert}, we calculate the attention probabilities of \cls to other tokens by averaging and normalizing the attention value in 12 attention heads in the last layers.
The attention distributions of the \method and its ablation variant are visualized in Figure~\ref{fig:attention_dist}.
We find that the tokens of \textit{writer} and \textit{novel} both have high attentions in the vanilla BERT-based concept extractor.
However, after adopting our knowledge-guided prompt, the attention probabilities of \textit{novel} is lower than before, and thus can help the model to reduce the spurious co-occurrence correlations derived from pre-trained knowledge.


\paragraph{What if other knowledge injection methods are adopted?}
We claim that the topics obtained from external KGs are better than the keyword-based topics from the text on guiding BERT to achieve our CE task. 
To justify it, we compare \method with another variant, namely \method$_{LDA}$, where the topics are the keywords obtained by running Latent Dirichlet Allocation (LDA)~\cite{2001Latent} over the abstracts of all entities.
Besides, we also compare \method with ERNIE~\cite{sun2019ernie}, which implicitly learns the knowledge of entities during pre-training. The detail about LDA and ERNIE is shown in Appendix~\ref{sec:LDA}.
The comparison results are listed in Table~\ref{tab:Knowledge}.
It shows that our design of the knowledge-guided prompt in \method exploits the value of external knowledge more thoroughly than the two remaining schemes, thus achieving better CE performance.

\setlength\tabcolsep{4pt}
\begin{table}[t]
\small
  \centering
 \begin{tabular}{lcccc}
 \toprule
 \textbf{Model}  & \textbf{NC \#} & \textbf{Prec.}   & \textbf{Recall$_R$}  & \textbf{F1$_R$} \\
 \midrule
 \rowcolor[gray]{0.95}\multicolumn{5}{c}{\textit{Trained on CN-DBpedia}} \\

 \method & \textbf{482}  & \textbf{94.20\%}   & \textbf{85.23\%} & \textbf{89.49\%} \\
 \method$_{LDA}$ & 308 & 93.08\%&82.13\%&87.26\% \\
 ERNIE & 302 & 93.86\% & 80.53\% & 86.69\% \\
 
 \midrule
 \rowcolor[gray]{0.95}\multicolumn{5}{c}{\textit{Trained on Probose}} \\

 \method & \textbf{752}&\textbf{88.69\%}&\textbf{80.85\%}&\textbf{84.59\%} \\
 \method$_{LDA}$ & 381&68.23\%&61.45\%&64.66\% \\
 ERNIE & 286 & 77.96\% & 46.13\% & 57.97\% \\

 \bottomrule
 \end{tabular}
  \caption{Concept extraction results with different knowledge utilization.}
  \label{tab:Knowledge}
\end{table}



\subsection{Applications}\label{subsec:Applications}

\setlength\tabcolsep{3pt}
\begin{table}[t]
    \small
    \centering
    \begin{tabular}{lccccc}
    \toprule
      \textbf{Model} & \textbf{TS \#} & \textbf{NC \#} & \textbf{Prec.} & \textbf{Recall$_R$} & \textbf{F1$_R$}  \\
    \midrule
    \method & 0 & 62 & 82.66\% & 48.44\% & 61.08\%\\ 
     \ \ \wo \textit{P} & 0 & 55 & 69.62\% & 42.97\% & 53.14\%\\  
    \cdashlinelr{1-6}
    \method & 300 & \textbf{107} & \textbf{82.95\%} & \textbf{83.59\%} & \textbf{83.27\%}\\ 
     \ \ \wo \textit{P} & 300 & 89 & 81.65\% & 69.53\% & 75.10\%\\ 
    \bottomrule
    \end{tabular}
    \caption{Human evaluation on 100 CE results for Meituan entities. TS \# is the number of training samples.}
    \label{tab:meituan}
\end{table}

\paragraph{KG Completion}
We run \method for all entities existing in CN-DBpedia to complement new concepts. 
\method extracts 7,623,111 new concepts for 6 million entities.
Thus, our framework can achieve a large-scale concept completion for existing KGs.

\paragraph{Domain Concept Acquisition}
We collect 117,489 Food \& Delight entities with their descriptive texts from Meituan~\footnote{http://www.meituan.com, a Chinese e-business platform.}, and explore two application approaches. 
The first is to directly apply \method, and the second is to randomly select 300 samples as a small training set to fine-tune \method. 
The results in Table \ref{tab:meituan} show that:
\begin{inparaenum}[\it 1)]
    \item The transfer ability of \method is greatly improved with the aid of prompts;
    \item \method can extract high-quality concepts in the new domain only with a small portion of training samples.
\end{inparaenum}
Furthermore, after running directly, \method extracts 81,800 new concepts with 82.66\% precision. 
Thus, our knowledge-guided prompt can significantly improve the transfer ability of PLMs on the domain CE task.

\section{Conclusion}
\label{sec:conclusion}
In this paper, we identify the concept bias in the PLM-based CE system and devise a Structural Causal Model to analyze the bias. 
To alleviate concept bias, we propose a novel CE framework with knowledge-guided prompting to alleviate spurious co-occurrence correlation between entities and biased concepts. 
We conduct extensive experiments to justify that our prompt-based learning framework can significantly mitigate bias and has an excellent performance in concept acquisition.

\section{Limitations}
Although we have proven that our work can significantly alleviate concept bias and extract high-quality and new concepts, it also has some limitations. 
In this section, we analyze three limitations and hope to advance future work.

\paragraph{Model Novelty}
Although \method can effectively mitigate the spurious co-occurrence correlations between entities and biased concepts, the proposed framework is not entirely novel.
The novelty of our work is to conduct the first thorough causal analysis that shows the spurious correlations between entities and biased concepts in the concept extraction task.
After defining the problem and SCM of concept extraction in $\mathsection$~\ref{subsec:Preliminary}, we propose a prompt-based approach to implement the interventions toward the SCM to elicit the unbiased knowledge from PLMs.
Previous work in language prompting mostly guides the PLMs with prompts but is unaware of the cause-effect relations in its task, which may hinder the effectiveness of prompts.
We hope our work can inspire future work to utilize language prompting from a causal perspective.

\paragraph{Topic Classification}
Although the topics obtained by clustering are mostly mutually exclusive, there are still cases where an entity can be classified into multiple topics. 
Therefore, considering only one topic for the entity excludes the correct concepts.

\paragraph{Threshold Selection}
We only reserve concepts with confidence scores bigger than the selection threshold ($\mathsection$~\ref{subsec:extractor}), which can hardly achieve a satisfactory balance of precision and recall.
If we select a relatively big threshold, we can get more accurate concepts but may lose some correct ones. 
If the recall is preferred, precision might be hurt.

We suggest that future work consider these three limitations to achieve better performance in the CE task.

\section*{Acknowledgement}
We would like to thank the anonymous reviewers for their valuable comments and suggestions for this work. This work is supported by the Chinese NSF Major Research Plan (No.92270121), Shanghai Science and Technology Innovation Action Plan (No.21511100401) and the Science and Technology Commission of Shanghai Municipality Grant (No. 22511105902).

\bibliography{anthology}

\clearpage
\appendix
\section{Theoretical Details of Causal Framework}\label{sec:theoretical_details}
\subsection{Preliminaries}
\paragraph{SCM} 
The Structural Causal Model (SCM) is associated with a graphical causal model to describe the relevant variables in a system and how they interact with each other.
An SCM $G=\{V,f\}$ consists of a set of nodes representing variables $V$, and a set of edges between the nodes as functions $f$ to describe the causal relations.
Figure~\ref{fig:SCM} shows the SCM that describes the PLM-based CE system. 
Here the input text $X$ serves as the \textit{treatment}, and the extracted concept span $S$ is the \textit{outcome}.
In our SCM, pre-trained knowledge $K$ is a cause of both $X$ and $S$, and thus $K$ is a \textit{confounder}.
A confounder can open \textit{backdoor paths} (\ie, $X \leftarrow K \rightarrow S$) and cause a spurious correlation between $X$ and $S$.
To control the confounding bias, intervention techniques with the do-operator can be applied to cut off backdoor paths.

\paragraph{Causal Intervention} 
To identify the true causal effects of $X \rightarrow S$, we can adopt the causal intervention to fix the input $X=x$ and removes the correlation between $X$ and its precedents, denoted as $do(X=x)$.
In this way, the true causal effects of $X \rightarrow S$ can be represented as $P(S=s|do(X=x))$.
The backdoor adjustment and the frontdoor adjustment are two operations to implement interventions and obtain $P(S=s|do(X=x))$.

Next, we will elaborate on the details of the two operations.

\subsection{The Backdoor Adjustment}
The backdoor adjustment is an essential tool for causal intervention. 
For our SCM, the pre-trained knowledge blocks the backdoor path between $X$ and $S$, then the causal effect of $X=x$ on $S$ can be calculated by:
\begin{align}\label{eq:ba}
	&P(S=s|do(X=x))\notag\\
	&=P_m(S=s|X=x)\notag\\
	&=\sum_{k} P_m(S=s|X=x,K=k)P_m(K=k)\notag\\
	&=\sum_{k} P(S=s|X=x,K=k)P(K=k),
\end{align}
where $P_m$ is the probability after applying the do-operator, and $P(K=k)$ needs to be estimated from data or priorly given.
However, it is intractable to observe the pre-training data and obtain the prior distribution of the pre-trained knowledge.
Therefore, the back adjustment is not applicable in our case.

\subsection{The Frontdoor Adjustment}
The frontdoor adjustment is a complementary approach to applying the intervention when we cannot identify any set of variables that obey the backdoor adjustment.

In our SCM, we aim to estimate the direct effect of $X$ on $S$, while being unable to directly measure pre-trained knowledge $K$.
Thus, we introduce a topic prompt $P$ as a mediator, and then the frontdoor adjustment can adopt a two-step do-operation to mitigate bias.
\paragraph{Step 1}
As illustrated in $\mathsection$~\ref{sec:SCM_analysis}, we first analyze the causal effect $X \rightarrow P$.
Since the collider, \ie, $X \rightarrow P \rightarrow S \leftarrow K$ blocks the association between $P$ and $K$, there is no backdoor path from $X$ to $P$. 
Thus we can obtain the conditional probability as (same as Eq.\ref{eq:step1}):
\begin{equation}\label{eq:step1_appendix}
	P(P=p|do(X=x))=P(P=p|X=x).
\end{equation}

To explain Step 1, we take an entity \textit{Louisa May Alcott} with her abstract as an example.
We can assign the topic \textit{person} as a prompt to make the PLM-based extractor alleviate spurious correlation between \textit{Louisa May Alcott} and \textit{novel}, and concentrate on extracting person-type concepts. 

\paragraph{Step 2}
In this step, we investigate the causal effect $P \rightarrow S$.
$P \leftarrow X \leftarrow K \rightarrow S$ contains a backdoor from $P$ to $S$. 
Since the data distribution of $X$ can be observed, we can block the backdoor path through $X$:
\begin{align}\label{eq:fa}
	&P(S=s|do(P=p))\notag\\
	&=\sum_{x} P(S=s|X=x,P=p)P(X=x),
\end{align}
where $P(X=x)$ can be obtained from the distribution of the input data, and $P(S=s|X=x,P=p)$ is the conditional probability of the extracted span given the abstract with a topic prompt, which can be estimated $P(S=s|X=x,P=p)$ by the concept extractor.

Now we can chain the two steps to obtain the causal effect $X \rightarrow S$:
\begin{align}\label{eq:final_appendix}
	&P(S|do(X))\notag\\
	&=\sum_{p} P(S|p,do(X))P(p|do(X))\notag\\
	&=\sum_{p} P(S|do(P),do(X))P(p|do(X))\notag\\
	&=\sum_{p} P(S|do(P))P(p|do(X)).
\end{align}

\section{Detailed Information about \method}

\setlength\tabcolsep{3pt}
\begin{table}[!hbt]
	\centering
	\small
	\begin{tabular}{ll}
		\toprule
		\textbf{Topic} &  \textbf{Corresponding Concept Examples} \\
		\midrule
		Person &  politicians, teachers, doctors\\
		
		Book & romance novels, art books, fantasy novels\\
		
		Location & towns, villages, scenic spots\\
		
		Film and TV & movies, animation, TV dramas\\
		
		Language & idioms, medical terms, cultural terms\\
		
		Game & electronic games, web games, mobile games\\
		
		Creature & plants, animals, bacteria\\
		
		Food & Indian cuisine, Japanese cuisine\\
		
		Website & government websites, corporate websites\\
		
		Music & singles, songs, pop music\\
		
		Software & application software, system software\\
		
		Folklore & social folklore, belief folklore\\
		
		Organization & companies, brands, universities\\
		
		History & ancient history, modern history\\
		
		Disease & digestive system disease, oral disease\\
		
		Technology & technology products, electrical appliances\\
		
		Medicine & Chinese medicine, Western medicine\\
		\bottomrule
	\end{tabular}
        \caption{17 typical topics and their corresponding concept examples.}
        \label{tb:topic}
\end{table}

\subsection{Identifying Topic for Each Entity}\label{sec:classifier}
The 17 typical topics and their corresponding concepts are listed in Table~\ref{tb:topic}.
We predict the topic of the entity as one of the 17 typical topics using a transformer encoder-based topic classifier.
We randomly fetch 40,000 entities together with their existing concepts in the KG. 
According to the concept clustering results, we can assign each topic to the entities. 
Specifically, we concatenate $E$ and $X$ as input to the classifier.
With multi-headed self-attention operation over the input token sequence, the classifier takes the final hidden state (vector) of a token $\cls$, i.e., $\mathbf{h}_{\cls}^{N_{L}}\in\mathbb{R}^{d''}$, to compute the topic probability distribution $P(T^i|E,X)\in\mathbb{R}^{17}$, where $N_{L}$ is the total number of layers and $d''$ is the vector dimension. 
Then, we identify the topic with the highest probability $T$ as the topic of $X$, which is calculated as follows,
\begin{align}\label{eq:classifer}
        \mathbf{H}^{0} &= \mathbf{EW}^{0}+\mathbf{B}^{0}, \\
		\mathbf{H}^{l} &= encoder(\mathbf{H}^{l-1}), 1\leq l \leq N_{L},\\
		P(T) &= \textrm{softmax}(\mathbf{h}_{\cls}^{N_{L}}\mathbf{W}^{L}), \\
		T &= \mathop{\arg\max}\limits_{i}\big(P(T^{i})\big), 1\leq i \leq 17
\end{align}
where $\mathbf{E}\in\mathbb{R}^{(|\mathbf{E}|+|\mathbf{X}|)\times d}$ is the random initial embedding matrix of all input tokens and $d$ is the embedding size. $\mathbf{H}^l\in\mathbb{R}^{(|\mathbf{E}|+|\mathbf{X}|)\times d''}$ is the hidden matrix of the $l$-th layer.
$\mathbf{h}_{\cls}^{N_{L}}$ is obtained from $\mathbf{H}^{N_{L}}$. 
Furthermore, $\mathbf{W}^{0} \in\mathbb{R}^{d \times d''}, \mathbf{B}^{0} \in\mathbb{R}^{(|\mathbf{E}|+|\mathbf{X}|) \times d''}$ and $\mathbf{W}^{L}\in\mathbb{R}^{d'' \times 17}$ are both training parameters. 

\begin{figure}[t]
	\centering
 \includegraphics[width=1.0\columnwidth]{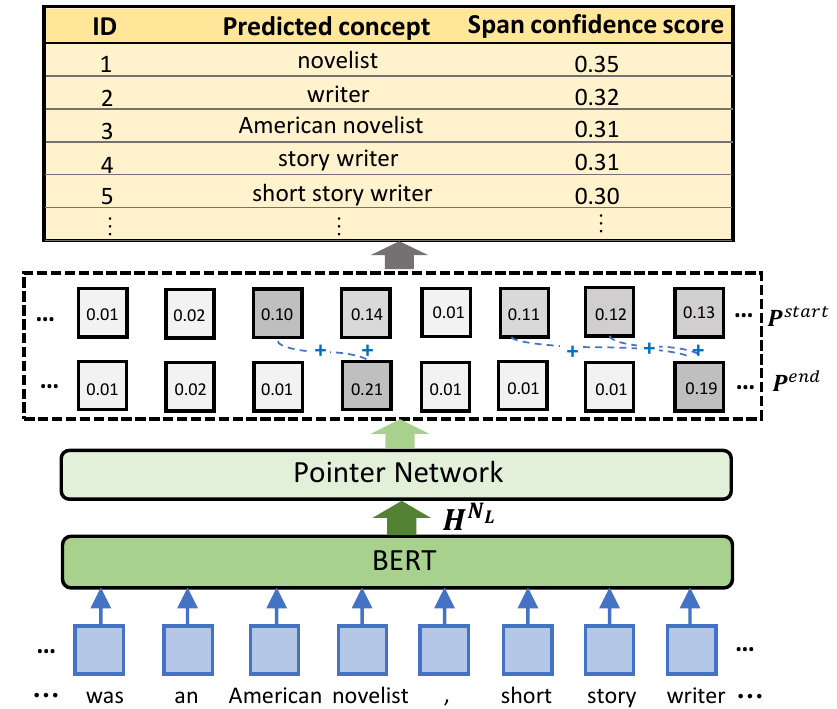} 
	\caption{An example to illustrate how to perform the pointer network.}
	\label{fig:PN}
\end{figure}

\subsection{An Example for Point Network}\label{sec:PN}
As mentioned in $\mathsection$~\ref{subsec:extractor}, we adopt a point network to achieve multi-grained concept extraction~\cite{yuan2021large}. 
The model generates a ranking list of candidate concepts (spans) along with their confidence scores, and outputs the concepts with confidence scores bigger than the selection threshold. 
Note that one span may be output repeatedly as the same subsequence of multiple extracted concepts through an appropriate selection threshold. 

For example, as shown in Figure~\ref{fig:PN}, \textit{writer} is extracted multiple times as the subsequence of three different granular concepts when the confidence score threshold is set to 0.30. 
Therefore, the point network enables our framework to extract multi-grained concepts.

\section{Experiment Detail}\label{sec:Baselines}
\subsection{Baselines}\label{sec:base}
We compare \method with seven baselines. 
Most of the compared models are the extraction models feasible for extraction tasks, including Named Entity Recognition (NER), Relation Extraction (RE), and Open Information Extraction (Open IE). 
 In addition, we also compare the pattern-based approach. 
However, we do not compare ontology extension models and generation models, since both do not meet our scenario. 
Since entity typing models cannot find new concepts, they are also excluded from our comparison. 
Please note that, except MRC-CE, other baselines applied in concept extraction cannot extract multi-grained concepts.
\begin{itemize}

\item \textbf{Hearst}~\cite{jiang2017metapad}:
With specific handwritten rules, this baseline can extract concepts from free texts. 
We design 5 Hearst patterns listed in Table~\ref{tb:pattern} where we translate the Chinese patterns for the Chinese dataset into English. 

\item  \textbf{FLAIR}~\cite{akbik2019flair}:
It is a novel NLP framework that combines different words and document embeddings to achieve excellent results. FLAIR can also be employed for concept extraction since it can extract spans from the text.

\item  \textbf{XLNet}~\cite{yang2020clinical}:
With the capability of modeling bi-directional contexts, this model can extract clinical concepts effectively.

\item  \textbf{KVMN}~\cite{nie2020improving}:
As a sequence labeling model, KVMN is proposed to handle NER by leveraging different types of syntactic information through the attentive ensemble.

\item  \textbf{XLM-R}~\cite{XLM-R,lange2022clin}:
It is a Transformer-based multilingual masked language model incorporating XLM~\cite{XLM} and RoBERTa~\cite{RoBERTa}, which has proven to be effective in extracting concepts.

\item  \textbf{BBF}~\cite{luo2021privacy}:
It is an advanced version of BERT built with Bi-LSTM and CRF. 
With optimal token embeddings, it can extract high-quality medical and clinical concepts.

\item \textbf{GACEN}~\cite{ijcai2021p200}:
The model incorporates topic information into feature representations and adopts a neural network to pre-train a soft matching module to capture semantically similar tokens.

\item  \textbf{MRC-CE}~\cite{yuan2021large}:
MRC-CE handles the concept extraction problem as a Machine Reading Comprehension (MRC) task built with an MRC model based on BERT. 
It can find abundant new concepts and handle the problem of concept overlap well with a pointer network.
\end{itemize}


\begin{table}[!htb]
	\centering
	\small
	\caption{The Hearst patterns used in the baseline.}  \label{tb:pattern} 
	\begin{tabular}{ll}
        \toprule
		\textbf{Dataset} & \textbf{Pattern}\\
        \midrule
		\multirow{5}{*}{\textbf{CN-DBpedia}}   & X is Y\\
		& X is one of Y  \\
		& X is a type/a of Y\\
		& X belongs to Y \\
		& Y is located/founded/ in...  \\
        \midrule
		\multirow{5}{*}{\textbf{Probase}}  &  X is a Y  that/which/who  \\
		& X is one of Y\\
		& X refers to Y   \\
		& X is a member/part/form... of Y  \\
		& As Y, X is ...  \\
    \bottomrule
	\end{tabular}
\end{table}

\subsection{Experiment Settings}\label{sec:set}
Our experiments are conducted on a workstation of dual GeForce GTX 1080 Ti with 32G memory and the environment of torch 1.7.1. 
We adopt a BERT-base with 12 layers and 12 self-attention heads as the topic classifier and concept extractor in \method. 
The training settings of our topic classifier are: d = 768, batch size = 16, learning rate = 3e-5, dropout rate = 0.1 and training epoch = 2. 
The training settings of our concept extractor are: d = 768, m = 30, batch size = 4, learning rate = 3e-5, dropout rate = 0.1 and training epoch = 2. 
The $\alpha$ in Eq.\ref{eq:loss1} is set to 0.3 and the selection threshold of candidate spans in the concept extractor is set to 0.12 based on our parameter tuning.

\subsection{Human Assessment}\label{sec:Human_Assessment}
Some extracted concepts do not exist in the KG, which cannot be automatically assessed. 
Therefore, we invite some volunteers to assess whether the extracted concepts are correct for the given entities. 
We denote an extracted concept as an \textit{EC} (existing concept) that has already existed in the KG for the given entity.
We denote an extracted concept as an \textit{NC} (new concept) represents a correct concept not existing in the KG for the given entity.
We employ four annotators in total to ensure the quality of the assessment. 
All annotators are native Chinese and proficient in English.
Each concept is labeled with 0, 1 or 2 by three annotators, where 0 means a wrong concept for the given entity, while 1 and 2 represent EC and NC, respectively. 
If the results from the three annotators are different, the fourth annotator will be hired for a final check.
We protect the privacy rights of the annotators and pay the annotators above the local minimum wage.

\subsection{Other Knowledge Injection Methods}\label{sec:LDA}
As we mentioned before, the topics of the knowledge-guided prompt come from external KGs, which are better than the keyword-based topics from the text on guiding BERT to achieve accurate concept extraction. 

To justify it, we compared \method with another variant, namely \method$_{LDA}$, where the topics are the keywords obtained by running Latent Dirichlet Allocation (LDA)~\cite{2001Latent} over all entities' abstracts. 
Specifically, the optimal number of LDA topic classes was also determined as 17 through our tuning study. 
For a given entity, its topic is identified as the keyword with the highest probability of its topic class. 
Besides, we also compared \method with ERNIE. ERNIE~\cite{sun2019ernie} adopts entity-level masking and phrase-level masking to learn language representation. 
During pre-training of ERNIE, all words of the same entity mentioned or phrase are masked. 
In this way, ERNIE can implicitly learn the prior knowledge of phrases and entities, such as relationships between entities and types of entities, and thus have better generalization and adaptability. 

The comparison results are listed in Table~\ref{tab:Knowledge}, which shows that our design of the knowledge-guided prompt in \method exploits external knowledge's value more thoroughly than the rest two schemes.

\label{sec:appendix}
\end{document}